\documentclass[runningheads]{llncs}
\usepackage{graphicx}
\usepackage{subcaption} 
\usepackage{tabularx} 
\usepackage[usenames,dvipsnames]{color}
\usepackage{authblk}
\usepackage{float}

\begin{document}
\title{Rice Plant Disease Detection and Diagnosis using Deep Convolutional Neural Networks and Multispectral Imaging}
%
\author{Yara Ali Alnaggar\textsuperscript{1} \and Ahmad Sebaq\textsuperscript{1} \and Karim Amer\textsuperscript{1} \and ElSayed Naeem\textsuperscript{2} \and Mohamed Elhelw\textsuperscript{1}}

\authorrunning{Y. et al.}
%

\institute{Center for Informatics Science, Nile University, Giza, Egypt\\
\email{\{y.ali,a.sebaq,k.amer,melhelw\}@nu.edu.eg} \and Rice Research Institute, Kafr ElSheikh, Egypt \\
\email{naeemlab72@yahoo.com}}

\maketitle
\begin{abstract}
Rice is considered a strategic crop in Egypt as it is regularly consumed in the Egyptian people's diet. Even though Egypt is the highest rice producer in Africa with a share of 6 million tons per year \cite{elbasiouny2020rice}, it still imports rice to satisfy its local needs due to production loss, especially due to rice disease. Rice blast disease is responsible for 30\% loss in rice production worldwide \cite{nalley2016economic}. Therefore, it is crucial to target limiting yield damage by detecting rice crops diseases in its early stages. This paper introduces  a public multispectral and RGB images dataset and a deep learning pipeline for rice plant disease detection using multi-modal data. The collected multispectral images consist of Red, Green and Near-Infrared channels and we show that using multispectral along with RGB channels as input archives a higher F1 accuracy compared to using RGB input only. 

\end{abstract}

\section{Introduction}
In Egypt, rice is important in Egyptian agriculture sector, as Egypt is the largest rice producer in Africa. The total area used for rice cultivation in Egypt is about 600 thousand ha or approximately 22\% of all cultivated area in Egypt during the summer.
As a result, it is critical to address the causes of rice production loss to minimize the gap between supply and consumption. Rice plant diseases contribute mostly to this loss, especially rice blast disease. According to \cite{nalley2016economic}, rice blast disease causes 30\% worldwide of the total loss of rice production. Thus, rice crops diseases detection, mainly rice blast disease, in the early stages can play a great role in restraining rice production loss.

Early detection of rice crops diseases is a challenging task. One of the main challenges of early detection of such disease is that it can be misclassified as the brown spot disease by less experienced agriculture extension officers (as both are fungal diseases and have similar appearances in their early stage) which can lead to wrong treatment. Given the current scarcity of experienced extension officers in the country, there is a pressing need and opportunity for utilising recent technological advances in imaging modalities and computer vision/artificial intelligence to help in early diagnosis of the rice blast disease. Recently, multispectral photography has been deployed in agricultural tasks such as precision agriculture \cite{candiago2015evaluating}, food safety evaluation \cite{QIN2013157}. Multispectral cameras could capture images in Red, Red-Edge, Green and Near-Infrared bands wavebands, which captures what the naked eye can’t see. Integrating the multispectral technology with deep learning approaches would improve crops diseases identification capability. However, it would be required to collect multispectral images in large numbers.

In this paper, we propose a public multispectral and RGB images dataset and a deep learning pipeline for rice plant disease detection. First, the dataset we present contains 3815 pairs of multispectral and RGB images for rice crop blast, brown spot and healthy leaves. Second, we developed a deep learning pipeline trained on our dataset which calculates the Normalised Difference Vegetation Index (NDVI) channel from the multispectral image channels and concatenates it along its RGB image channels. We show that using NDVI+RGB as input archives a higher F1 score by 1\% compared to using RGB input only. 

\section{Literature Review}
Deep learning has emerged to tackle problems in different tasks and fields. Nowadays, it is being adopted to solve the challenge of crop disease identification. For example, Mohanty et al. \cite{mohanty2016using} trained a deep learning model to classify plant crop type and its disease based on images. Furthermore, \cite{amara2017deep} proposed a deep learning-based approach for banana leaf diseases classification. 

Furthermore, multispectral sensors have proven its capability as a new modality to detect crop fields issues and diseases. Some approaches use multispectral images for disease detection and quantification. Cui et al. \cite{cui2010image} developed an image processing-based method for quantitatively detecting soybean rust severity using multi-spectral images. Also, \cite{zhang2018detection} utilize digital and multispectral images captured using quadrotor unmanned aerial vehicles (UAV) to collect high-spatial resolution imagery data to detect the ShB disease in rice.

After the reliable and outstanding results deep learning models could achieve on rgb images, some approaches were developed to use deep learning on multispectral images, especially of crops and plants. \cite{osorio2020deep} proposed a deep learning-based approach for weed detection in lettuce crops trained on multispectral images. In addition, Ampatzidis et al. \cite{ampatzidis2019uav} collects multispectral images of citrus fields using UVA for crop phenotyping and deploys a deep learning detection model to identify trees. 

\section{Methodology}

\subsection{Hardware Components}
\begin{figure}[h]
        \centering
    \includegraphics[width=0.5\textwidth]{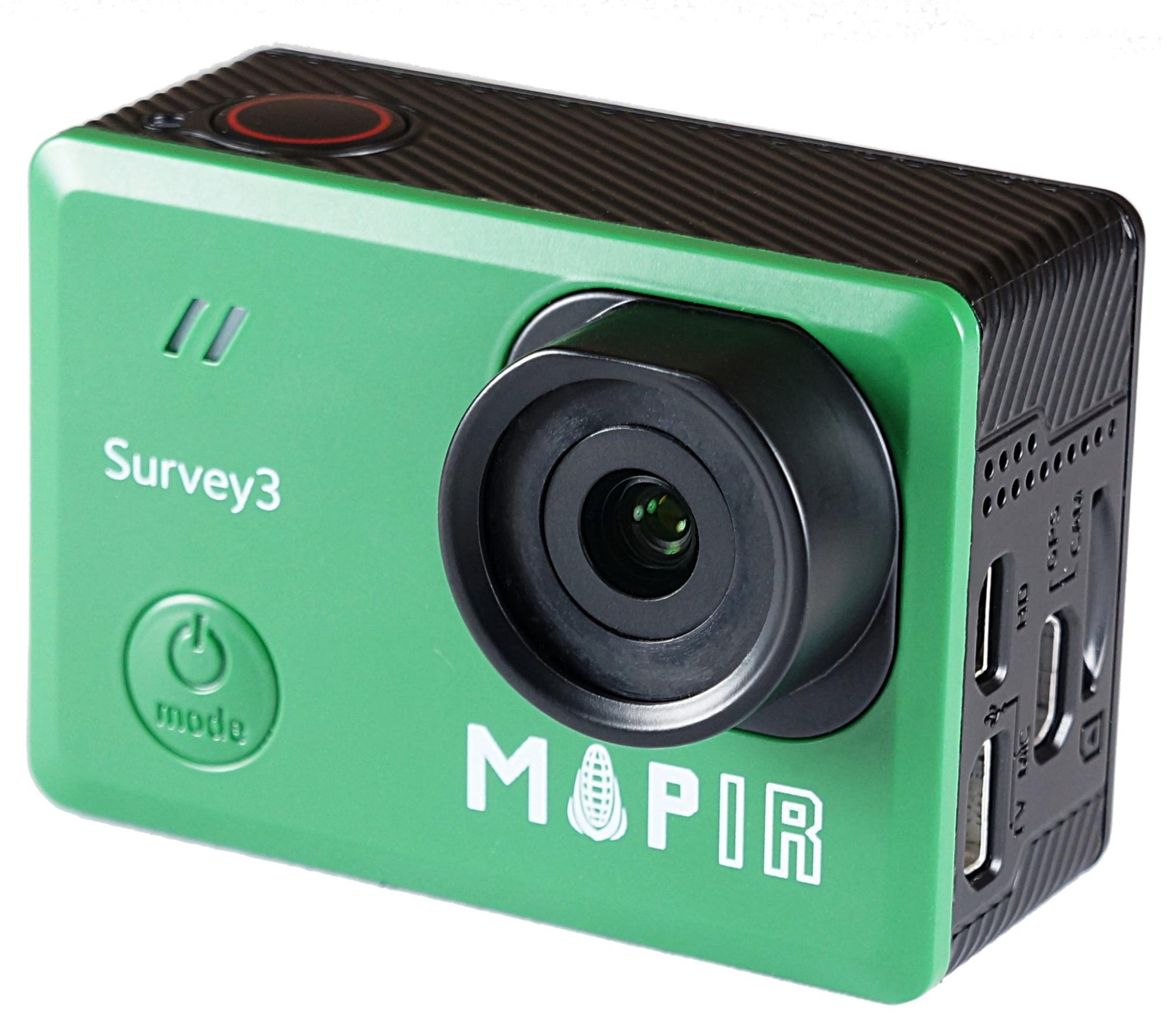}
    \caption{MAPIR Survey3N Camera.}
    \label{fig:mapir_camera}
\end{figure}

We used a MAPIR Survey3N camera, shown in Figure \ref{fig:mapir_camera}  to collect our dataset. This camera model captures ground‐level multispectral images of red, green and NIR channels. It was chosen in favour of its convenient cost and easy integration with smartphones.. In addition, we used the Samsung Galaxy M51 mobile phone camera to capture RGB images, paired with the MAPIR camera.

\begin{figure}[h]
     \centering
     \begin{subfigure}[h]{0.4\textwidth}
         \centering
         \includegraphics[width=\textwidth]{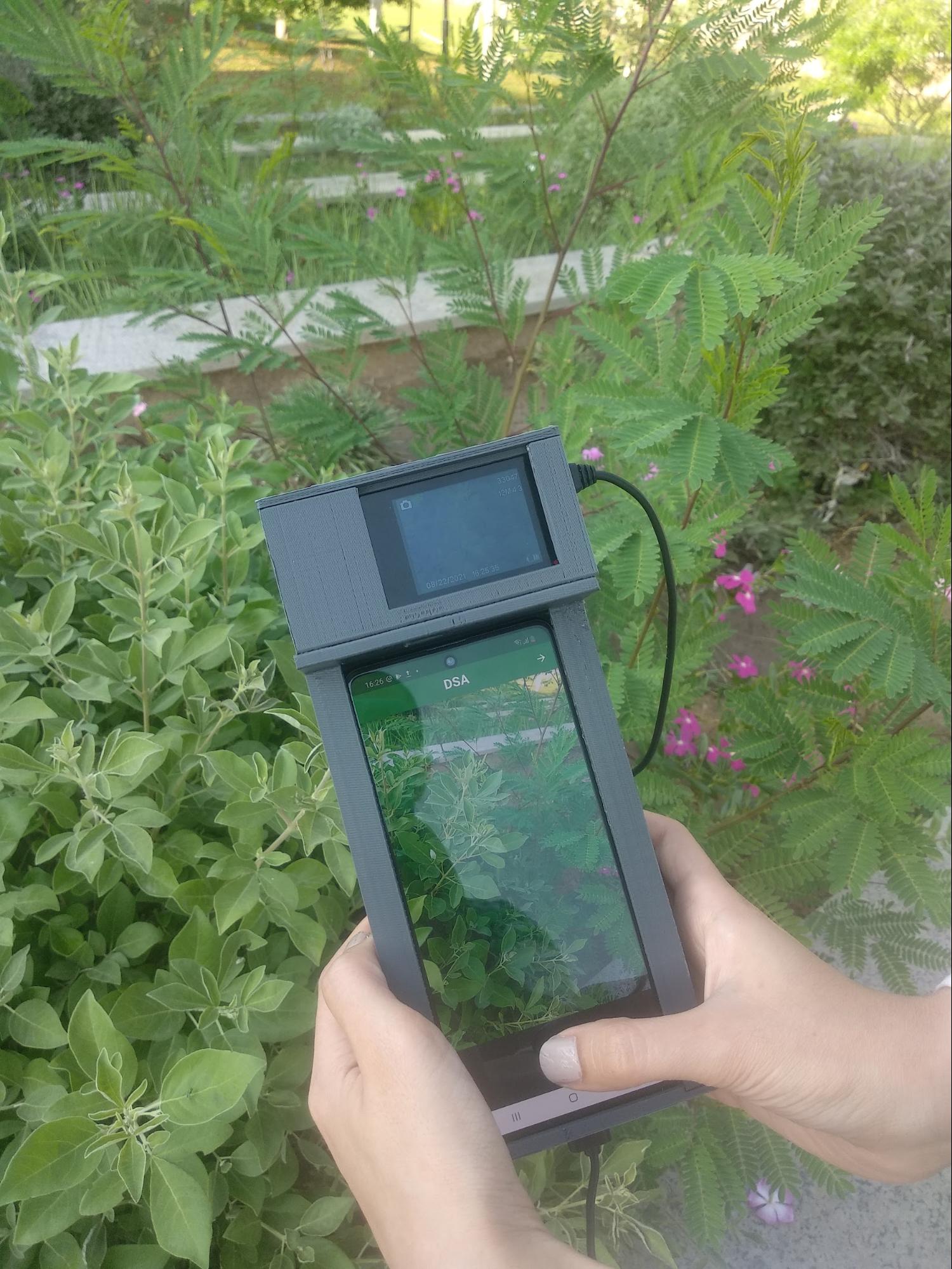}
     \end{subfigure}
     \begin{subfigure}[h]{0.4\textwidth}
         \centering
         \includegraphics[width=\textwidth]{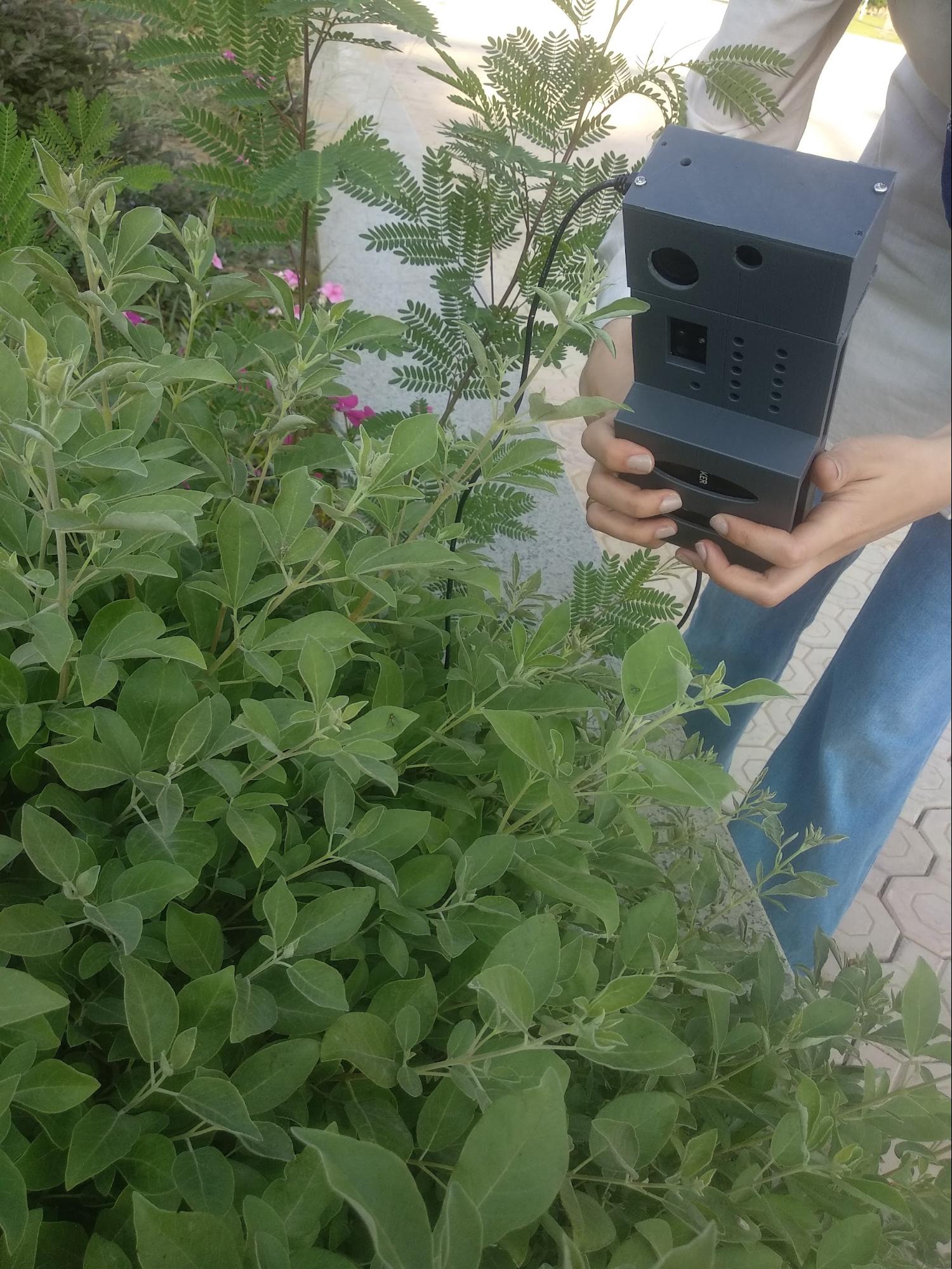}
     \end{subfigure}
        \caption{Holder gadget.}
        \label{fig:app_live}
\end{figure}

We Designed a holder gadget to combine the mobile phone, MAPIR camera and a power bank in a single tool, as seen in Figure \ref{fig:app_live}, to facilitate the data acquisition operation for the officers. It was designed using SolidWorks software and manufactured by a 3D printer. 

\subsection{Data Collection Mobile Application}
An android frontend application was also developed to enable the officers who collect the dataset to control the multispectral and the smartphone cameras for capturing dual RGNIR/RGB images simultaneously while providing features such as image labelling, imaging session management, and Geo‐tagging. The mobile application is developed with Flutter and uses Firebase real‐time database to store and synchronise the captured data including photos and metadata. Furthermore, Hive local storage database is used within the application to maintain a local backup of the data.

\subsection{Analytics Engine Module}
\begin{figure}[h]
        \centering
    \includegraphics[width=0.7\textwidth]{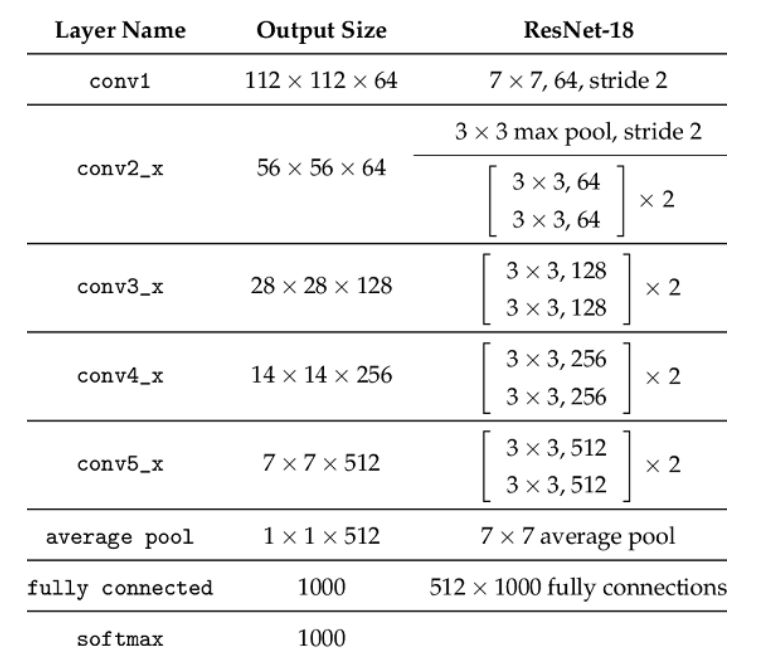}
    \caption{ResNet18 original architecture}
    \label{fig:original_resnet}
\end{figure}

Our engine is based on ResNet18 \cite{he2016deep} architecture which consists of 18 layers and it utilize the power of residual network, see Figure \ref{fig:original_resnet}, residual network help us avoid the vanishing gradient problem.

We can see how layers are configured in the ResNet-18 architecture. The architecture starts with a convolution layer with 7x7 kernel size and stride of 2. Next we begin with the skip connection. The input from here is added to the output that is achieved by 3x3 max pool layer and two convolution layers with kernel size 3x3, 64 kernels each. This is the first residual block.

The output of this residual block is added to the output of two convolution layers with kernel size 3x3 and 128 such filters. This constituted the second residual block. Then the third residual block involves the output of the second block through skip connection and the output of two convolution layers with filter size 3x3 and 256 such filters. The fourth and final residual block involves output of third block through skip connections and output of two convolution layers with same filter size of 3x3 and 512 such filters.

Finally, average pooling is applied on the output of the final residual block and received feature map is given to the fully connected layers followed by softmax function to receive the final output. \\

The vanishing gradient is a problem which happens when training artificial neural networks that involved gradient based learning and backpropagation. We use gradients to update the weights in a network. But sometimes what happens is that the gradient becomes very small, effectively preventing the weights to be updated. This leads to network to stop training. To solve such problem, residual neural networks are used.

\begin{figure}[h]
        \centering
    \includegraphics[width=0.7\textwidth]{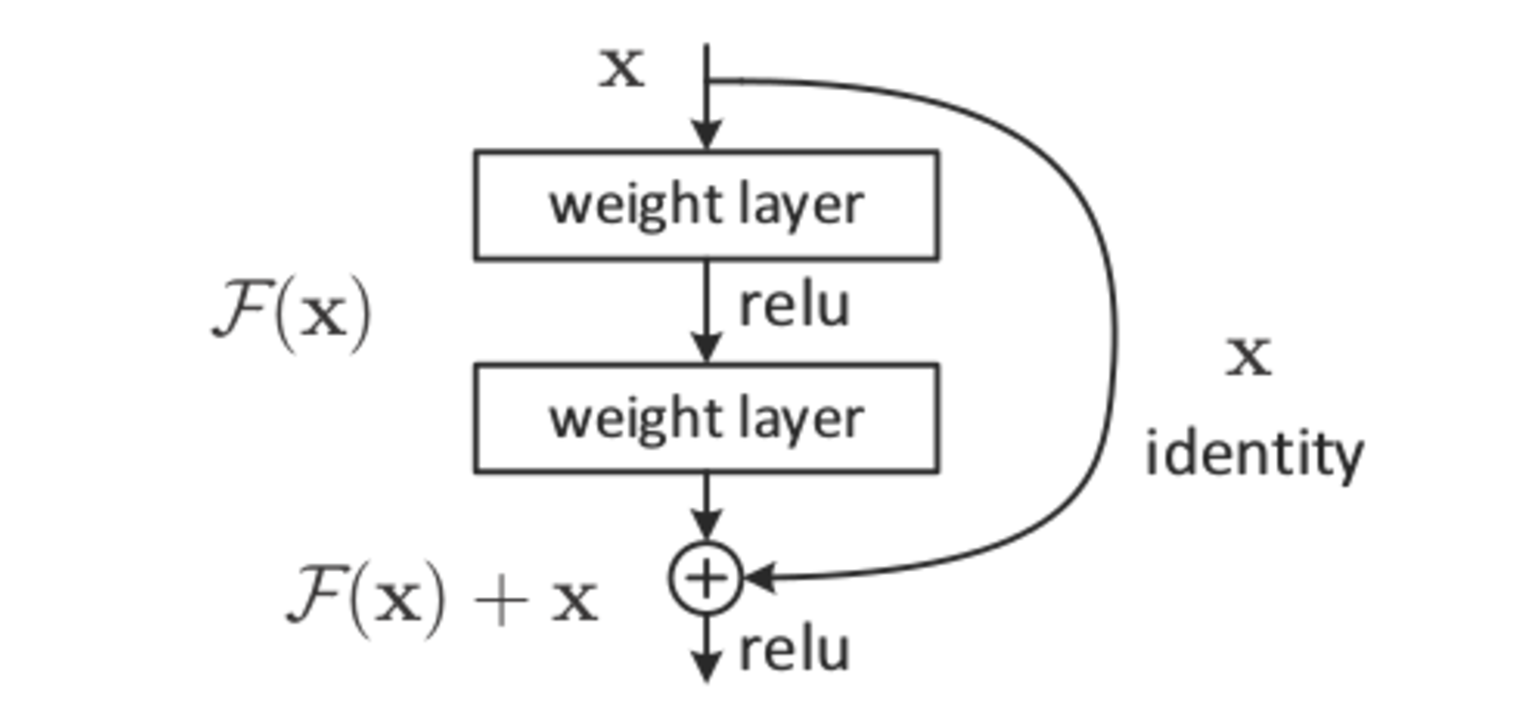}
    \caption{Residual block}
    \label{fig:resnet}
\end{figure}

Residual neural networks are the type of neural network that applies identity mapping. What this means is that the input to some layer is passed directly or as a shortcut to some other layer.
If $x$ is the input, in our case its an image or a feature map, and $F(x)$ is the output from the layer, then the output of the residual block can be given as $F(x) + x$ as shown in Figure \ref{fig:resnet}.

We changed the input shape to be 256x256 instead of 224x244, also we replaced the last layer in the original architecture with a fully connected layer where the output size was modified to three to accommodate our task labels. 

\section{Experimental Evaluation}

\subsection{Dataset}
\begin{figure}[h]
        \centering
    \includegraphics[width=0.7\textwidth]{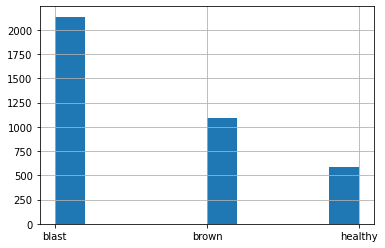}
    \caption{Collected dataset distribution.}
    \label{fig:dataset_classes}
\end{figure}

\begin{figure}[h]
     \centering
     \begin{subfigure}[h]{0.45\textwidth}
         \centering
         \includegraphics[width=\textwidth]{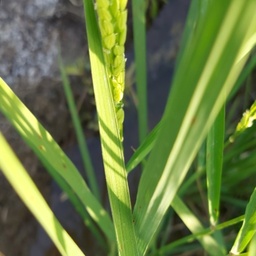}
     \end{subfigure}
     \begin{subfigure}[h]{0.45\textwidth}
         \centering
         \includegraphics[width=\textwidth]{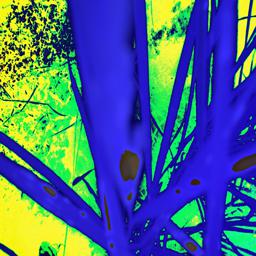}
     \end{subfigure}
        \caption{On the left is the RGB image and on the right is its R-G-NIR pair.}
        \label{fig:rgb_rgnir}
\end{figure}

\begin{figure}[h!]
    \centering
        \begin{subfigure}[b]{\textwidth}
            \centering
            \includegraphics[width=0.4\linewidth]{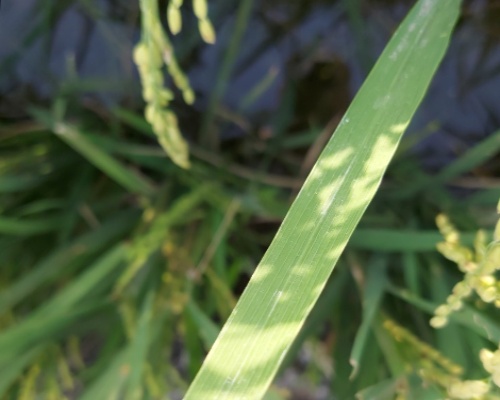}
            \includegraphics[width=0.425\linewidth]{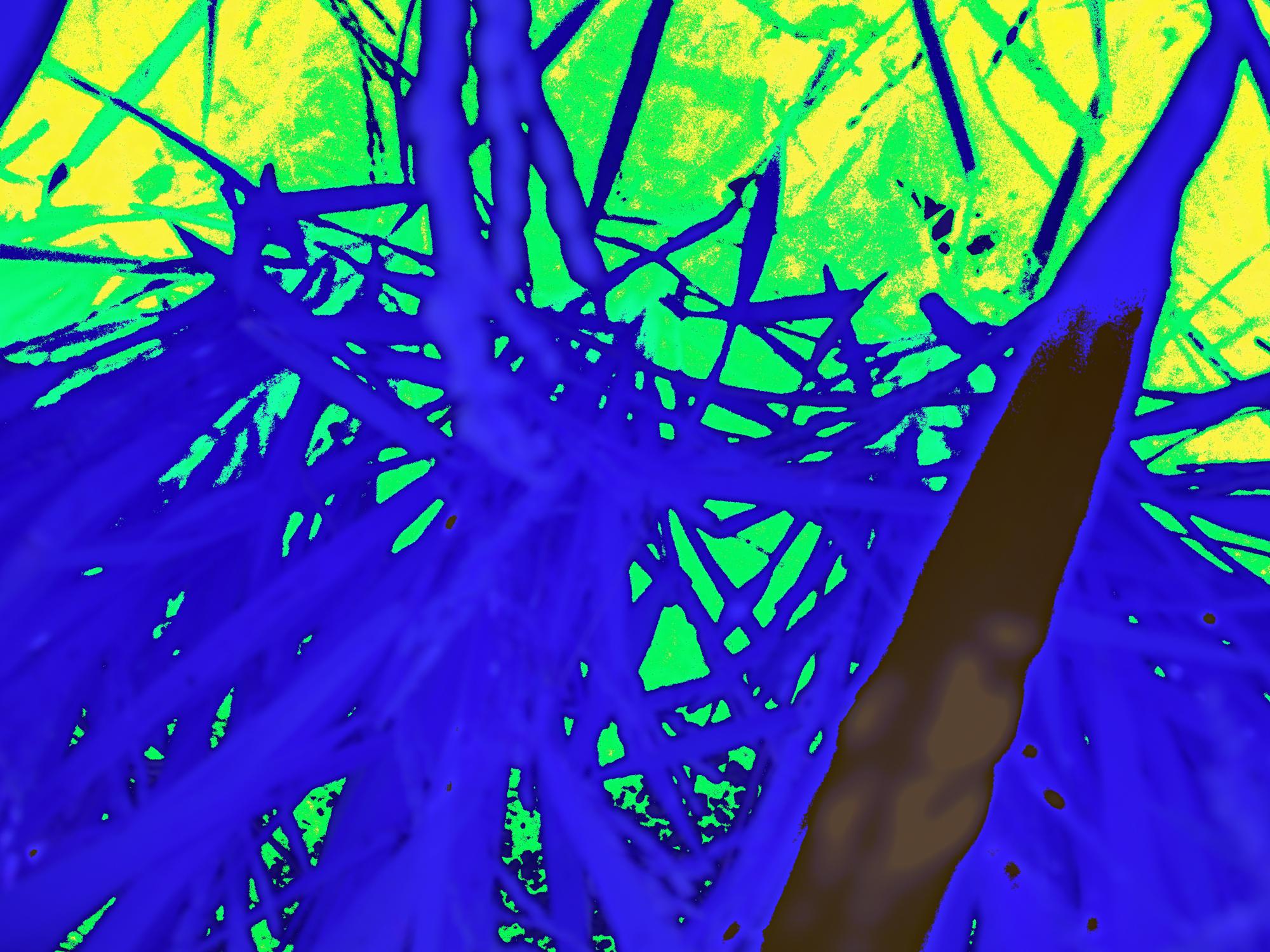}
        \caption{}
     \end{subfigure}
    \vskip\baselineskip    
    \centering
        \begin{subfigure}[b]{\textwidth}
            \centering
            \includegraphics[width=0.4\linewidth]{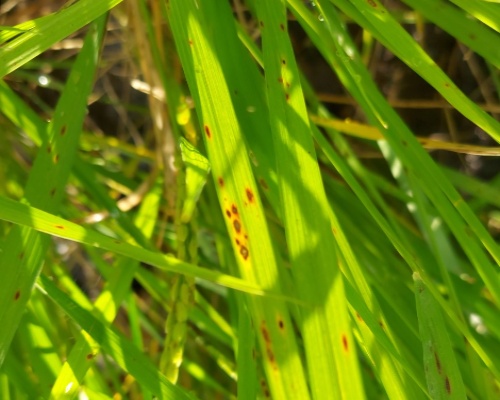}
            \includegraphics[width=0.425\linewidth]{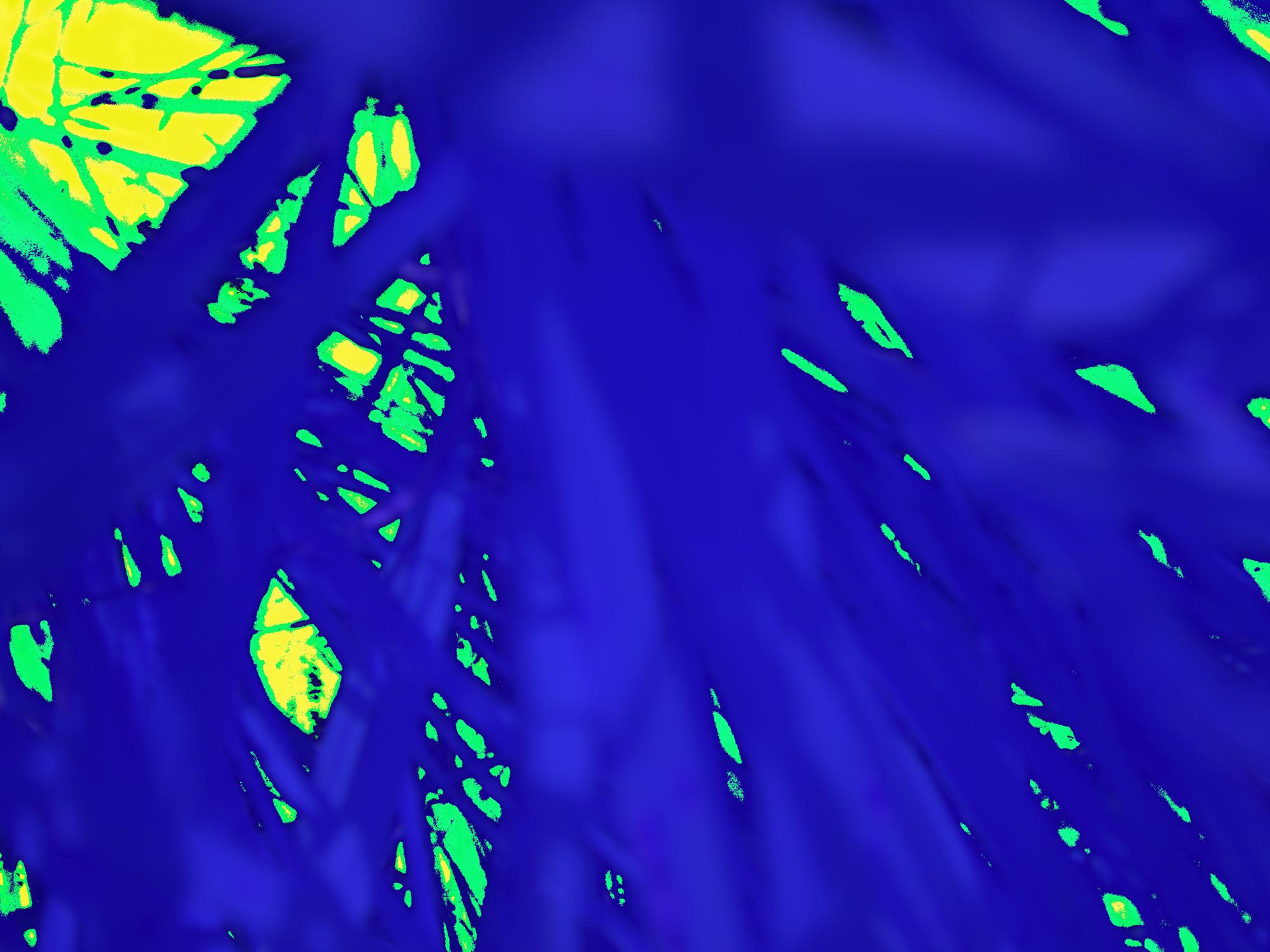}
        \caption{}
     \end{subfigure}
    \vskip\baselineskip    
    \centering
        \begin{subfigure}[b]{\textwidth}
            \centering
            \includegraphics[width=0.4\linewidth]{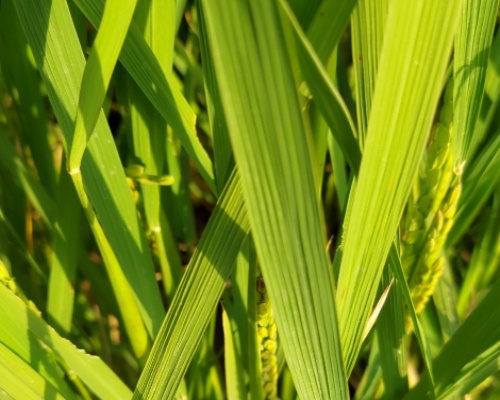}
            \includegraphics[width=0.425\linewidth]{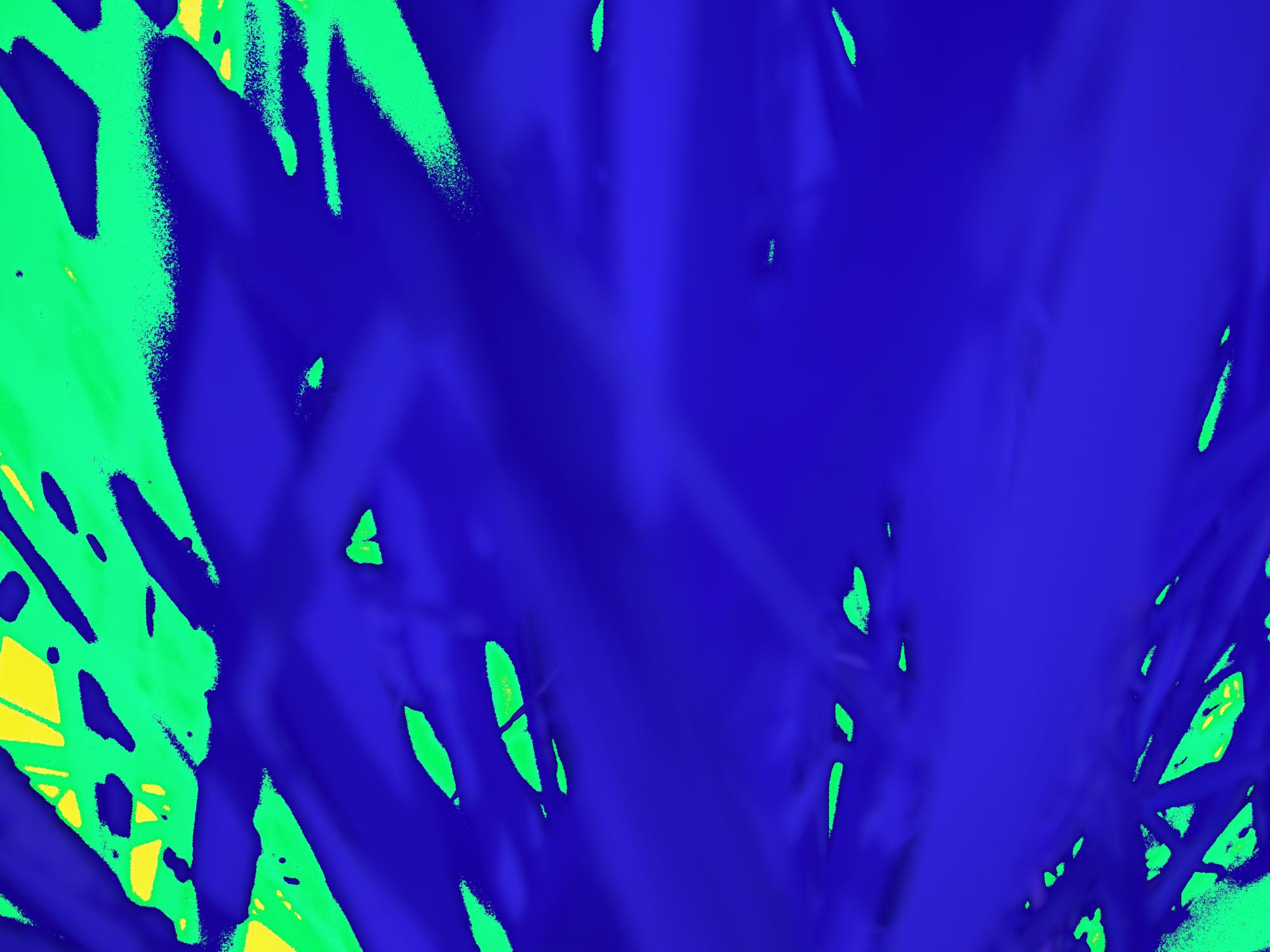}
        \caption{}
     \end{subfigure}
    
    \caption{(a) Blast class sample. (b) Brown spot class sample. (c) Healthy class sample.}
    \label{fig:classes_samples}
\end{figure}

We have collected 3815 samples of rice crops of three labels: blast disease, brown spot disease and healthy leaves distributed, shown in Figure \ref{fig:dataset_classes}, as the following: 2135, 1095 and 585, respectively. Each sample is composed of a pair of (RGB) and (R-G-NIR) images as seen in Figure \ref{fig:rgb_rgnir}, which were captured simultaneously.  Figure \ref{fig:classes_samples} shows samples of the three classes in our dataset. 

\subsection{Training Configuration}
In this section, we explain our pipeline for training data preparation and preprocessing. Also, we mention our deep learning models training configuration for loss functions and hyperparameters.

\subsubsection{Data Preparation}

\begin{figure}[h]
     \centering
     \begin{subfigure}[h]{0.25\textwidth}
         \includegraphics[width=\textwidth]{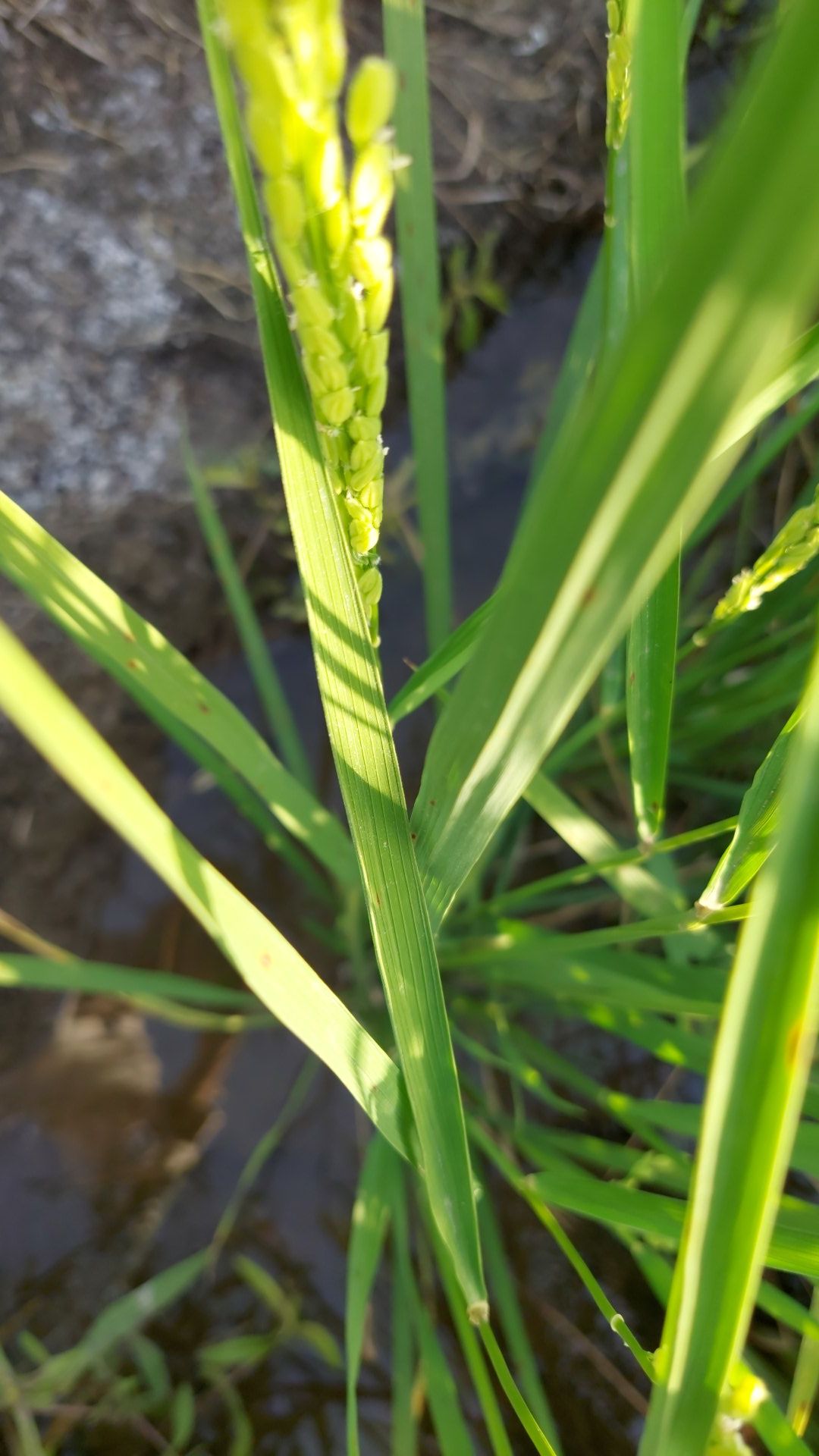}
     \end{subfigure}
     \begin{subfigure}[h]{0.45\textwidth}
         \includegraphics[width=\textwidth]{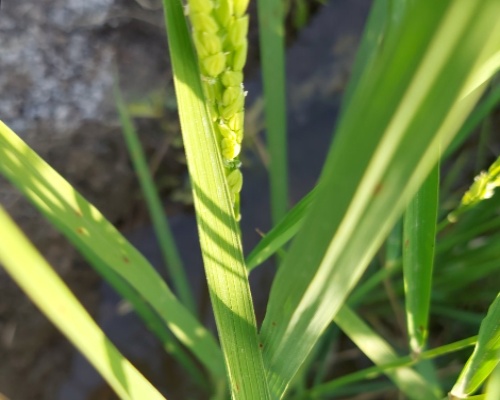}
     \end{subfigure}
        \caption{On the left is an RGB image before calibration and on the right is after registration.}
        \label{fig:register}
\end{figure}

\paragraph{RGB images registration}
Since the image sample of our collected dataset consists of a pair of RGB and R-G-NIR images, the two images are expected to have a similar field of view. However, the phone and MAPIR camera have different field of view parameters that the mapir camera has a 41° FOV compared to the phone camera with 123° FOV. As a result, we register the rgb image to the r-g-nir image using the OpenCV library. The registration task starts by applying an ORB detector over the two images to extract 10K features. Next, we use a brute force with Hamming distance matcher between the two images extracted features. Based on the calculated distances for the matches, we sort them descendingly and drop the last 10\%.  Finally, the homography matrix is calculated using the matched points in the two images to be applied over the RGB images. Figure \ref{fig:register} shows an RGB image before and after registration.

\begin{figure}[h]
        \centering
    \includegraphics[width=0.7\textwidth]{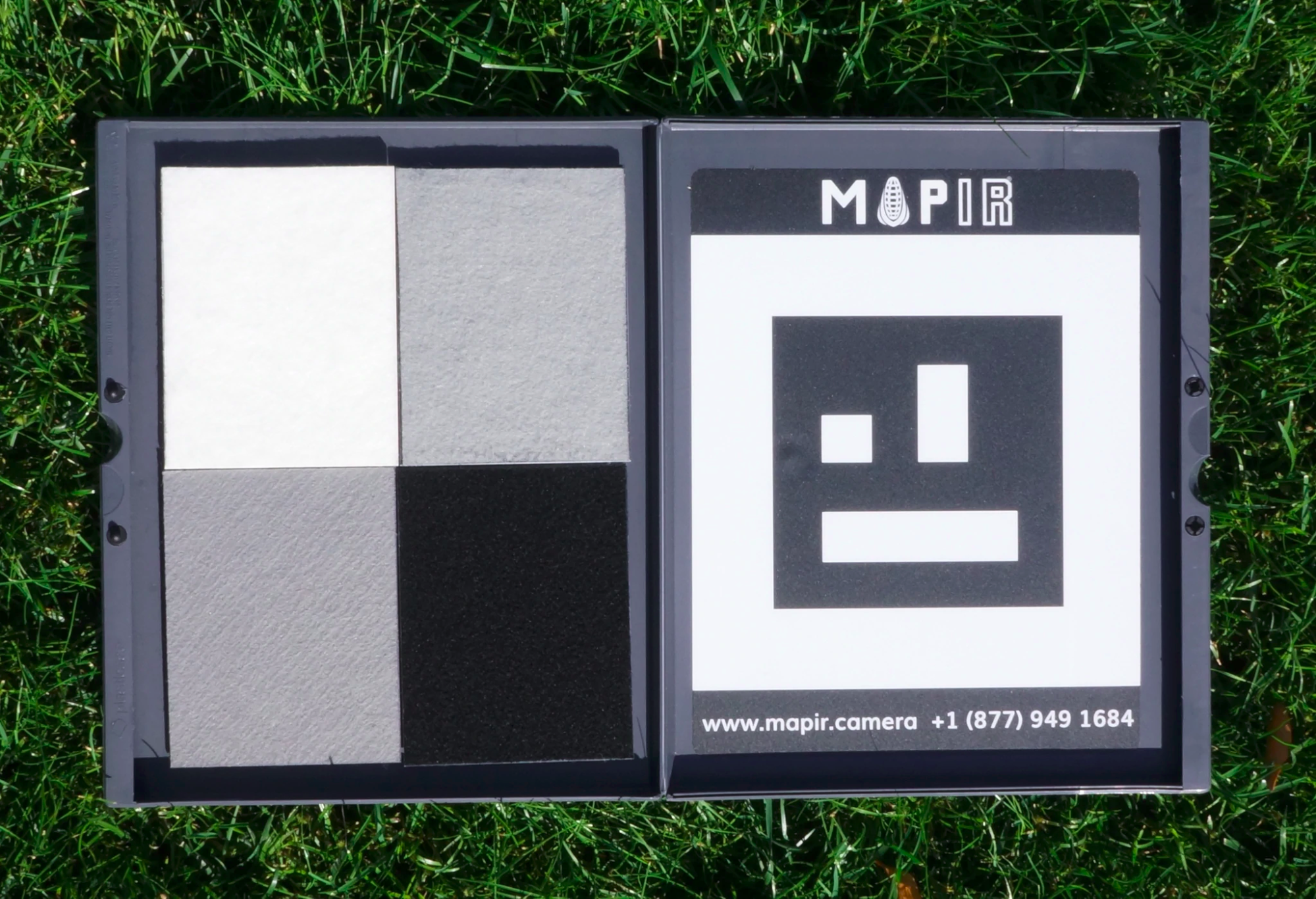}
    \caption{MAPIR Camera Reflectance Calibration Ground Target board.}
    \label{fig:kit}
\end{figure}

\paragraph{MAPIR camera calibration}
The MAPIR camera sensor captures the reflected light which lies in the Wavelengths in the Visible and Near Infrared spectrum from about 400-1100n and saves the percentage of reflectance. After this step, calibration of each pixel is applied to ensure that it is correct. This calibration is performed before every round of images captured using the MAPIR Camera Reflectance Calibration Ground Target board, which consists of 4 targets with known reflectance values, as shown in Figure \ref{fig:kit}. 

\paragraph{Models training configuration}
We trained our models for 50 epochs with a batch size of 16 using Adam optimizer and Cosine Annealing with restart scheduler \cite{loshchilov2016sgdr} with cycle length 10 epochs and learning rate of 0.05. For the loss function, we used a weighted cross entropy to mitigate the imbalance of the training dataset. Images were resized to dimension 256 x 256.

\subsubsection{Results}
For training the deep learning model using RGB and R-G-NIR pairs, we generate a NDVI channel, using Equation \ref{eq:ndvi}, and concatenate it to the RGB image. Our study shows that incorporating the NDVI channel improves the model capability to classify the rice crops diseases. Our model could achieve a F1 score with 5-kFold of 84.9\% when using RGB+NDVI as input compared to using only RGB image which could obtain a F1 score of 83.9\%. Detailed results are presented in Table 1.

\begin{equation}
    NDVI = \frac{NIR-Red}{NIR+Red}
    \label{eq:ndvi}
\end{equation}

\begin{table}
\caption{F1 score over our collected dataset achieved by using RGB as input versus RGB+NDVI.}\label{tab1}
\centering
\begin{tabularx}{\textwidth}{|X|X|X|}
\hline
Class &  RGB & RGB+NDVI\\
\hline
Blast   &  89.64\% & 90.02\% \\
Spot    &  82.64\% & 83.26\%\\
Healthy &  79.08\% & 81.54\%\\
\hline
\end{tabularx}
\end{table}

\section{Conclusion}
We presented our public dataset and deep learning pipeline for rice plant disease detection. We showed that employing multispectral imagery with RGB improves the model capability of disease identification by 1\% compared to using solely RGB imagery. We believe using a larger number of images for training would enhance current results also considering a larger number of images when using a deeper model this will result in better results. In addition, more investigation on how to fuse multispectral imagery with RGB for training could be applied, for example we can calculate NDVI from the blue channel instead of the red this may also boost the model performance.
\subsubsection{Acknowledgements.
}
The authors would like to acknowledge the support received from Data Science Africa (DSA) which made this work possible.

%
%
%
\bibliographystyle{splncs04}

\bibliography{ref}
\end{document}